\documentclass[10pt,twocolumn,letterpaper]{article}

\usepackage[pagenumbers]{cvpr} 

\usepackage[dvipsnames,table,xcdraw]{xcolor}
\definecolor{cvprblue}{rgb}{0.21,0.49,0.74}
\definecolor{iccvblue}{rgb}{0.21,0.49,0.74}
\usepackage[pagebackref,breaklinks,colorlinks,citecolor=cvprblue]{hyperref}

\usepackage{graphicx}
\usepackage{caption}
\usepackage{arydshln}
\usepackage{tikz}
\usepackage{wrapfig}
\usepackage{float}
\usepackage{multirow}
\usepackage{pifont}
\usepackage{makecell}
\newcommand{\cmark}{\ding{51}}%

\newcommand{\figref}[1]{Fig.~\ref{#1}}
\newcommand{\tabref}[1]{Tab.~\ref{#1}}

\newcommand{\secref}[1]{Sec.~\ref{#1}}
\newcommand{\myPara}[1]{\vspace{6pt}\noindent\textbf{#1}}

\def\MyMthd{LLaVA-Octopus}

\newcommand*\samethanks[1][\value{footnote}]{\footnotemark[#1]}

\newcommand{\tablestyle}[2]{\setlength{\tabcolsep}{#1}\renewcommand{\arraystretch}{#2}\centering\small}

\def\paperID{344} 
\def\confName{CVPR}
\def\confYear{2025}

\title{\MyMthd{}: Unlocking Instruction-Driven \\Adaptive Projector Fusion  for Video Understanding}

\author{
Boyuan Sun$^{1,2}$\thanks{Equal contribution.}\quad
Jiaxing Zhao$^{2}$\samethanks \quad
Xiang Chen$^{1}$\quad
Xihan Wei$^1$\quad
Qibin Hou$^2$\thanks{Corresponding author.} \\
{\normalsize{$^1$VCIP, CS, Nankai University}} \quad
{\normalsize{$^2$Tongyi Group, Alibaba}}
\\
{\tt{\small{\{zjx244036,xchen.cx,xihan.wxh\}@alibaba-inc.com,}}}\\
{\tt{\small{boyuansun@mail.nankai.edu.cn,}}} \\
\textbf{\normalsize{\url{https://github.com/Jiaxing-star/LLaVA-Octopus}}}\vspace{-3mm}}

\begin{document}
\maketitle
\begin{tikzpicture}[remember picture,overlay,shift={(current page.north west)}]
    \node[anchor=north west,xshift=3cm,yshift=-3.4cm]{\includegraphics[width=1.5cm]{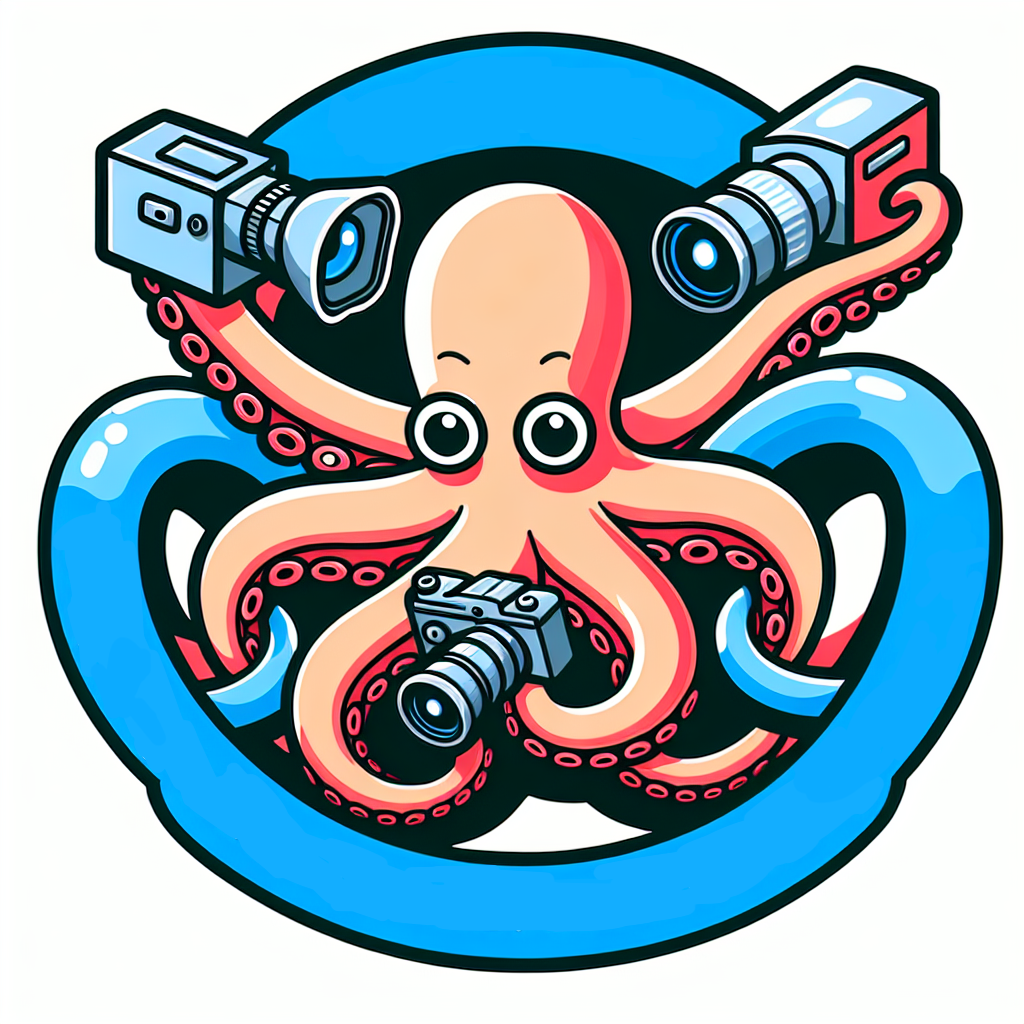}};
\end{tikzpicture}

\begin{abstract}
In this paper, we introduce \MyMthd{}, a novel video multimodal large language model. \MyMthd{} adaptively weights features from different visual projectors based on user instructions, enabling us to leverage the complementary strengths of each projector. We observe that different visual projectors exhibit distinct characteristics when handling specific tasks. For instance, some projectors excel at capturing static details, while others are more effective at processing temporal information, and some are better suited for tasks requiring temporal coherence.
By dynamically adjusting feature weights according to user instructions, \MyMthd{} dynamically selects and combines the most suitable features, significantly enhancing the model's performance in multimodal tasks.
Experimental results demonstrate that \MyMthd{} achieves excellent performance across multiple benchmarks, especially in tasks such as video question answering, long video understanding, and comprehensive multi-choices benchmarks, highlighting its broad application potential.

\end{abstract}    
\section{Introduction}
\label{sec:intro}
In recent years, the rapid advancement of multimodal large language models (MLLMs) \cite{gpt4v, zhao2025humanomni, geminiteam2024geminifamilyhighlycapable, alayrac2022flamingo,chen2024expanding,ye2023mplugowl,bai2023qwen,chen2023internvl,sun2025llava,jin2026geoagent} has led to significant progress in leveraging large language models~\cite{brown2020language,ouyang2022training,chatgpt,achiam2023gpt,vicuna2023,touvron2023llama2,jiang2024mixtral, guo2025deepseek,liu2024deepseek, qwen2.5} for image understanding.
However, human-computer interaction based solely on images is insufficient for many application scenarios, as most real-world interactions occur in video form.
The primary challenge in video understanding lies in managing temporal dynamics~\cite{damonlpsg2024videollama2}, as models must capture and interpret actions and events that evolve over time. 
Semantic understanding presents another major obstacle, as videos contain not only objects and actions but also complex semantic elements, such as character intentions and emotional expressions. Furthermore, the inherent complexity of video data, combined with the scarcity of high-quality annotated data, results in substantial computational costs and limits the model's learning capabilities. These factors make video understanding a more complex task than image understanding, attracting widespread research interest.

\begin{figure}
    \centering
    \vspace{-4mm}
    \includegraphics[width=1\linewidth]{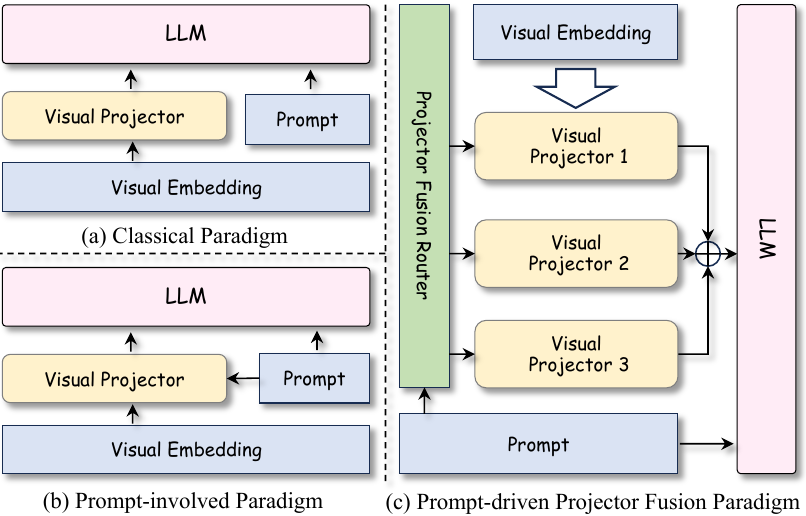}
    \vspace{-0.7cm}
    \caption{\textbf{Comparison of Different MLLM Paradigms.} In the classical paradigm, user instructions are fed into the LLM solely as text tokens. While the instruction-involved paradigm facilitates interaction between instructions and visual features, it is constrained by a single projector. Our proposed instruction-driven projector fusion paradigm designs a projector fusion router, which dynamically adjusts the weights of different types of visual projectors based on user instructions to produce the fused visual tokens.
    \vspace{-0.1cm}}
    \label{fig:teaser} 
    \vspace{-0.5cm}
\end{figure}

\begin{figure*}
    \centering
    \includegraphics[width=1\linewidth]{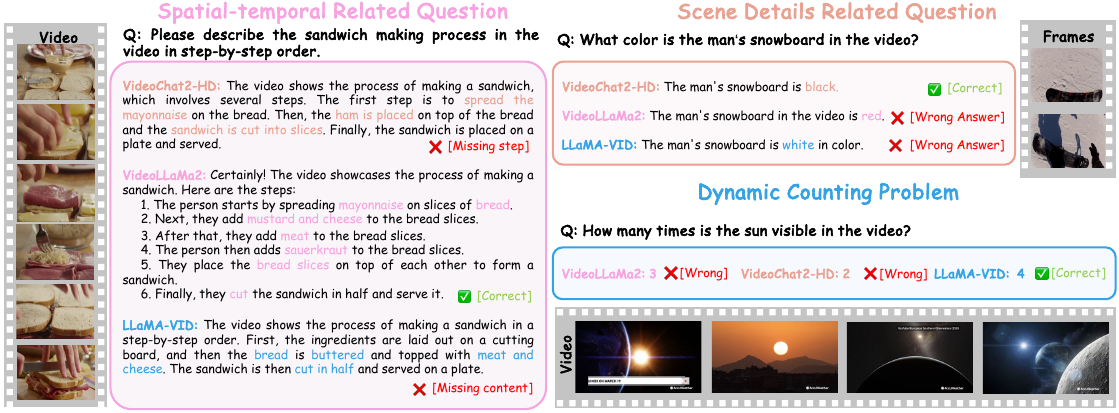}
    \vspace{-0.5cm}
    \caption{\textbf{Comparisons of three representative methods under different video understanding scenarios.} VideoChat2-HD~\cite{li2024mvbenchcomprehensivemultimodalvideo} uses image-based projector while VideoLLaMa2~\cite{damonlpsg2024videollama2} and LLaMA-VID~\cite{li2024llamavid} use spatial-temporal projector and token-compress projector, respectively. The results indicate that different visual projectors perform well in their appropriate domains while exhibiting poorer performance in other scenarios. More examples will be provided in the supplementary materials.}
    \label{fig: compare} 
    \vspace{-15pt}
\end{figure*}

As shown in Fig.~\ref{fig:teaser}, a typical video MLLM~\cite{maaz2023video,lin2023video,liu2024world, zhao2025facial, yang2025humanomniv2} consists of a visual encoder for feature extraction, a text encoder for textual representation, a visual projector to map visual features into a compatible space, and a large language model (LLM) decoder to generate contextually relevant text based on the combined representations.
Among them, the visual projector is crucial as it bridges the visual encoder and LLM, enabling visual understanding by mapping visual features into a space compatible with LLMs. 
Therefore, designing an appropriate visual projector for LLMs is a central focus in many MLLM works.

However, due to the varying video understanding scenarios that different MLLMs are designed to address, the projectors tailored for them exhibit distinct forms and characteristics.
In Fig.~\ref{fig: compare}, we present three representative video understanding tasks, offering an intuitive illustration of the characteristics of three typical approaches that employ different specifically designed visual projectors. Each approach demonstrates unique advantages within its specialized domain.
Therefore, we further categorize the visual projectors employed by these approaches into three types: image-based projectors, spatial-temporal projectors, and token-compress projectors.

The first type~\cite{li2024mvbenchcomprehensivemultimodalvideo} independently processes each frame and concatenates the results as visual tokens for LLM, offering an advantage in the comprehension of scene detail.
The second type~\cite{damonlpsg2024videollama2} utilizes a dedicated spatial-temporal module to capture inter-frame relationships, demonstrating strong performance on spatial-temporal related tasks.
However, due to efficiency constraints and limitations of LLMs, these two projectors often require frame sampling~\cite{li2024llava,damonlpsg2024videollama2,zhang2024llavanextvideo} before video input, resulting in the loss of many intermediate frames.
The third type~\cite{li2024llamavid} attempts to tackle this issue by compressing and reducing the number of tokens per frame, enabling the model to handle more frames and proving more effective for tasks requiring temporal coherence, such as counting problems.
Although projectors designed for specific tasks perform well in their domains of expertise, they struggle to handle complex video scenarios and diverse user instructions.
In addition, some methods use the instruction-involved paradigm shown in Fig.~\ref{fig:teaser} to emphasize the interaction between user instructions and visual features.
However, these approaches are limited by their reliance on a single type of projector and tend to fail to handle scenarios outside the projector's strengths.

Inspired by the aforementioned observations, we propose the instruction-driven projector fusion paradigm as shown in Fig.~\ref{fig:teaser}(c) and a model called \MyMthd{}.
This model introduces an instruction-driven adaptive router that integrates the strengths of different visual projectors based on user instructions. 
\MyMthd{} is able to adaptively adjust the feature weights of various visual projectors according to user instructions, thereby capitalizing on the complementary advantages of each projector. 
By dynamically combining the most appropriate features guided by user instructions, \MyMthd{} substantially enhances the model's performance in multiple video understanding tasks.
In \secref{sec:exp}, we conduct extensive ablation studies to demonstrate the feasibility of our proposed model. The results show that our model, \MyMthd{}, achieves state-of-the-art (SOTA) performance on most benchmarks and comparable performance on some benchmarks.

\section{Related Work}
\begin{figure*}[t]
    \vspace{-8mm}
    \centering
    \includegraphics[width=0.95\linewidth]{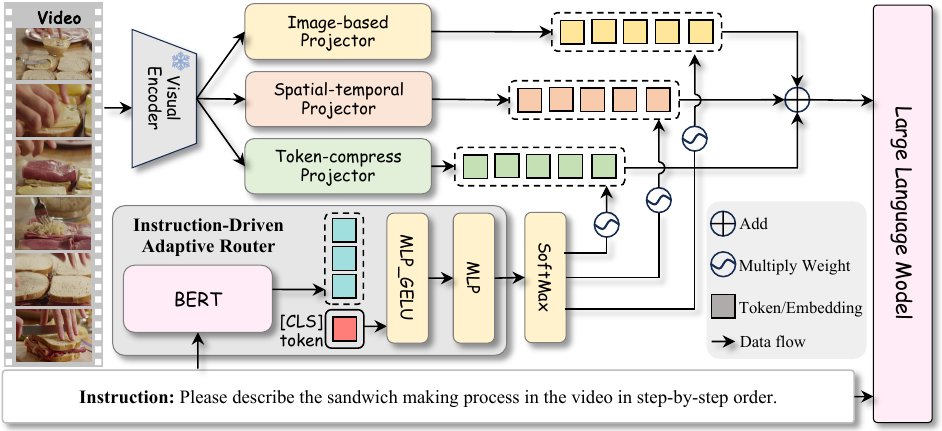}
    \vspace{-0.2cm}
    \caption{\textbf{Pipeline of the proposed \MyMthd{} model.} Our \MyMthd{} proposes an instruction-driven adaptive projector that involves three types of visual projectors to enhance the model's ability in multimodal tasks.}
    \label{fig: method} 
    \vspace{-15pt}
\end{figure*}

\subsection{Multimodal Large Language Model}
Currently, multimodal large language models can be categorized into community models and proprietary models. Proprietary models \cite{gpt4v, chatgpt, openai2023gpt4, Claude2024, geminiteam2024geminifamilyhighlycapable} often achieve better performance but are not open-sourced. 
Meanwhile, community models~\cite{ li2024llava, liu2023llava, liu2023llava1.5, zhang2024llavanext-video,li2024llavaone,li2024llavanext-ablations, sun2025depth, gao2023llama, damonlpsg2023videollama, damonlpsg2024videollama2,li2024llamavid}, which have seen rapid performance improvements, are garnering increasing attention due to their open-source nature, including model architecture, weights, and even training data.
LLaVA~\cite{liu2023llava} was the first to combine the powerful capabilities of LLMs with visual encoders like CLIP, enabling it to understand multimodal instructions and take actions accordingly, thus achieving comprehensive understanding and processing of visual and linguistic inputs. 
LLaVA1.5~\cite{liu2023llava1.5} encodes different types of data into vectors of the same dimension, allowing for the handling of more modalities. LLaVA-Next~\cite{li2024llavanext-strong,zhang2024llavanext-video} focuses more on processing video data, while LLaVA-OneVision~\cite{li2024llavaone} proposes a unified model capable of handling single images, multiple images, videos, audio, and other modalities simultaneously.

Based on the ideas of LLaVA, several variant series have emerged, such as the mPLUG-owl series. mPLUG-owl~\cite{ye2023mplug} introduces a new paradigm for training large language models through modularity, and the latest version, mPLUG-owl3~\cite{ye2024mplugowl3longimagesequenceunderstanding}, can even understand 2-hour movie videos.
BLIP-2~\cite{li2023blip2} uses Q-Former~\cite{zhang2023vision} to connect the visual and linguistic modalities. In BLIP-3~\cite{xue2024xgenmmblip3familyopen}, Q-Former is replaced by more scalable visual token samplers, such as perceptual resamplers.
We observe that numerous methods have explored various visual projectors. However, to the best of our knowledge, we are the first to classify these projectors and analyze their complementarity.
\subsection{Projector for Video MLLMs}
As described in \secref{sec:intro}, the specific designed visual projectors are crucial for LMMs.
We categorize them into three categories and select a representative method from each category to illustrate their strengths in \figref{fig: compare}.

\myPara{Image-based projector} refers to a projector that extracts features from every frame of the input video.
Considering the success of simple projectors such as linear projection~\cite{liu2024visual, chen2023minigptv2, liu2023llava, chen2023shikra} and cross-attention~\cite{Qwen-VL, wang2023cogvlm,ye2023mplugowl2} in Image LLMs, many Video LLMs~\cite{lin2023video, Qwen2-VL, li2023videochat,Maaz2023VideoChatGPT, ataallah2024minigpt4, li2024llavaone} directly adopt similar schemes as image-based projectors.
Besides, some more complex image-based projectors, such as Q-Former~\cite{zhu2023minigpt4, alayrac2022flamingo, dai2023instructblip, li2023blip2}, have also found applications in video MLMMs~\cite{damonlpsg2023videollama, li2024mvbenchcomprehensivemultimodalvideo}.
The image-based projector can capture detailed information within individual frames, thereby leading to superior performance in tasks related to scene details.
 However, limited to 
 the high computational cost
 and the absence of temporal modeling, the image-based projector faces challenges dealing with temporal related task.

\myPara{Spatial-temporal projector} aims to consider the relationships between video frames and attempt to reduce the number of visual tokens.
VideoLLaMa2~\cite{damonlpsg2024videollama2} introduces 3D convolution as the Spatial-Temporal Convolution Connector for spatial-temporal aggregation.
PLLaVA~\cite{xu2024pllava} integrates pooling strategies in both temporal and spatial dimensions.
VideoLLaMB~\cite{videollamb} designs recurrent memory bridge layers to preserve crucial visual information and semantic coherence.
Those spatial-temporal projectors provide significant advantages in handling spatial-temporal related question. 
However, the fusion of spatial and temporal information may lead to a loss of detailed image perception.

\myPara{Token-compress projector} is designed for enhancing the model's capacity to handle more input frames.
As a typical approach, LLaMa-VID~\cite{li2024llamavid} attempt to tackle the computation and memory challenges by compressing visual features.
BLIP-3-Video~\cite{blip3video-xgenmmvid} integrate adaptive pooling strategies to compress visual tokens.
LongVA~\cite{zhang2024longcontexttransferlanguage}, on the other hand, addresses the issue by expanding the capacity of LLMs, increasing the number of tokens they can process.
Some approaches~\cite{wang2024lifelongmemoryleveragingllmsanswering, zhang2023simple, wang2024videoagentlongformvideounderstanding} also consider agent-based techniques to convert visual inputs into textual descriptions.
Despite the token-compress projector's ability to increase the number of frames supported by LLMs and excel at handling videos with rapidly changing content, the compression of tokens per frame limits the perception of scene details and temporal information.

\section{Method}
In this section, we first introduce the motivation of \MyMthd{}  then describe its architecture, the detailed training process, and the implementation specifics.

\subsection{Motivation}
As discussed in Sec. 1, each type of visual projector excels in specific domains tailored to different user instructions. However, in practical scenarios, complex and multifaceted user instructions frequently transcend the boundaries of a single task, leading to unsatisfactory user experiences. Motivated by this, we propose a video MLLM that can handle various scenarios based on user instructions.

To achieve this, we first selected some widely adopted projectors in MLLMs  as candidates. Specifically, we chose the basic MLP2x\_GELU as the image-based projector $E_{img}$, the STC module from VideoLLaMA2~\cite{damonlpsg2024videollama2} as the spatial-temporal projector $E_{stc}$, and LLaMA-VID's~\cite{li2024llamavid} token-compress projector $E_{com}$. 
This selection ensures a comprehensive coverage of the diverse requirements posed by user instructions. 
%
Then, we design the instruction-driven adaptive router and build our \MyMthd{} upon it.

\subsection{\MyMthd{}}
In Fig.~\ref{fig: method}, we present a detailed architecture diagram of \MyMthd{}. \MyMthd{} primarily consists of four key components: a visual encoder, a series of visual projectors $E = \{E_{img}, E_{stc}, E_{com}\}$, an instruction-driven adaptive router $R$, and a large language model decoder. 
Among these components, the Instruction-Driven Adaptive Router is the core innovation of \MyMthd{}.

\myPara{Instruction-Driven Adaptive Router.}
%
For text instructions input $x_t$, we first use BERT~\cite{devlin2018bert} to encode the instructions, generating textual features of the user instructions. 
%
We focus on the [CLS] token output by BERT, which can effectively represent the semantics of the instruction, providing a solid foundation for subsequent weight generation.


Then, we leverage two multi-layer perceptrons (MLPs) to capture high-level semantic information from the instruction and generate $R(x_t)$ as the output of the instruction-driven adaptive router $R$. 
The first MLP\_GELU layer takes the [CLS] token as input and transforms it into the intermediate feature representation. 
%
%
The second MLP takes the intermediate feature representation as input and adjusts the output dimension to match the number of projectors, yielding $R(x_t) \in \mathbb{R}^3$.
This enables us to further process  $R(x_t)$ into a set of weights, where the relative magnitude of its values reflects the degree of alignment between the user instruction and each type of projector, and consequently, is utilized as the gate value for fusing multiple visual projector embeddings.

\myPara{Multiple Visual Projectors Embedding.} 
For the video input, we first use the visual encoder to obtain the visual embedding $ x_v$.  To ensure that the features obtained from the three different types of projectors are consistent in terms of token numbers, we make the following adjustments to the MLP, STC, and LLaMA-VID projectors.

First, for the image-based projector $E_{img}$, the original setting extracts 8 video frames, resulting in a token count of $14 \times 14 \times 8+8=1576$. To align the token counts, we remove the separators between each image, reducing the token count to 1568.
%
For spatial-temporal projector $E_{stc}$, the original setting results in a token count of $13 \times 13 \times 4=676$ for 8 video frames. To ensure token consistency, we modify the sampler parameters in the STC module. Specifically, we use a stride of (2, 2, 2) and (1, 2, 2), with padding of (1, 1, 1). These modifications ensure that the STC projector produces a token count of 1568. 
%
Finally, for token-compress projector $E_{com}$, we use 128 frames to represent the video. For each frame, we use 6 context tokens and 6 content tokens. To ensure token consistency, we add a separator token every 4 frames. Specifically, the number of tokens for every 4 frames is 49, and the total token count for 128 frames is $49 \times 32 = 1568$.

Through the above adjustments, we align the visual token counts from different projectors. Thus, we  can gather the output of each projector as the multiple visual projectors embedding $E(x_v)$:
\begin{equation}
    E(x_v) =\{E_{img}(x_v), E_{stc}(x_v), E_{com}(x_v)\}.
\end{equation}
Then, equipped with $R(x_t)$, we are able to dynamically combine the multiple visual projectors embedding $E(x_v)$ based on user text instructions.

\myPara{Projetcors Fusion.}
Given the output of the instruction-driven adaptive router $R(x_t) \in \mathbb{R}^3$, the gate-value for each projector can be obtained by:
\begin{equation}
    p_i(x_t) = \frac{e^{R(x_t)_i}}{\sum_{j}^{3} e^{R(x_t)_j}}.
\end{equation}
Then, with the set of multiple visual projectors embedding $E(x_v)$, we can calculate the final visual embedding $\mathcal{E}$ with: 
\begin{equation}
    \mathcal{E} = \sum_{i=1}^{3} p_i(x_t)\cdot E_{i}(x_v).
    \label{eq:feature_combination}
\end{equation}
After the fusion of multiple visual projectors embeddings, the large language model decoder takes the fused visual embedding $\mathcal{E}$ along with the text instruction $x_t$ and give the final prediction.

%
%

\begin{figure}[t]
    \centering
    \includegraphics[width=1\linewidth]{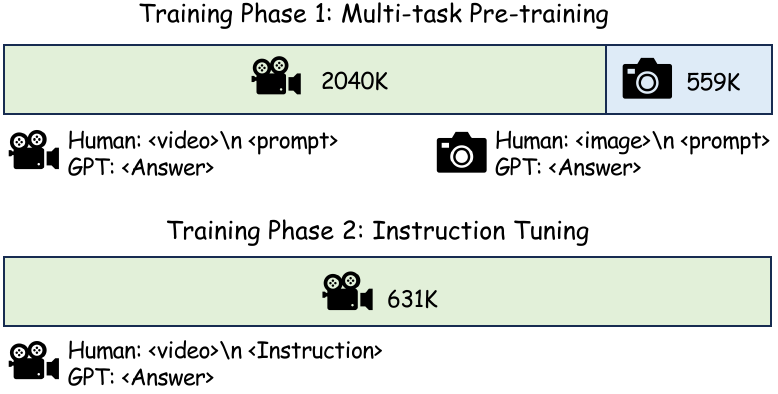}
    \caption{\textbf{Multimodal Data Distribution and Data Format.} $<$image$>$ and $<$video$>$ represent visual tokens from image and video data, respectively.}
    \label{fig: data format} 
\end{figure}

\begin{table}[t]
    \centering
    \small
    \setlength{\abovecaptionskip}{2pt}
    \definecolor{lightlightgray}{gray}{0.8}
    \tablestyle{7pt}{1.0}
    \scalebox{0.8}{
    \begin{tabular}{llccc}
        \toprule
        \textbf{Modality} &  \textbf{Dataset} & \textbf{Original}  & \textbf{Used} & \textbf{Ratio (\%)}\\
        \midrule
        &\multicolumn{4}{c}{\textbf{Multi-task Pre-training Stage}}\\
        \multirow{2}{*}{\textbf{Image-Text}} & CC-3M~\cite{sharma2018conceptual} & 3M & 558K & 18.6\% \\
                   & RealWorldQA~\cite{grok1.5} & 0.77K & 0.77K & 100\% \\
        \midrule
        \multirow{8}{*}{\textbf{Video-Text}} & WebVid-10M~\cite{Bain21} & 10M & 702K & 7.02\% \\
                   & CLVERER~\cite{yi2019clevrer} & 300K & 224K & 74.8\%\\
                   & NEXT-QA~\cite{xiao2021next} & 52K & 39K & 75.2\% \\
                   & Youcook2~\cite{zhou2018towards} & 2K & 1.79K & 89.5\% \\
                   & Charades~\cite{sigurdsson2016hollywood} & 27.8K & 19.7K &70.7\%\\
                   & Charades-Ego~\cite{sigurdsson2018charades} & 66.5K & 14.0K & 21.1\%\\
                   & TGIF~\cite{li2016tgif} & 120K & 120K & 100\% \\
                   & ShareGPT4Video~\cite{chen2024sharegpt4video} & 4.8M & 902K & 18.8\%\\
                   
        \midrule
        &\multicolumn{4}{c}{\textbf{Instruction Tuning}}\\
        \textbf{Hybrid} & Oryx~\cite{liu2024oryx} & 1.2M & 631K & 52.6\% \\
        \bottomrule
    \end{tabular}
    }
    \caption{Data Statistics of Multi-task Pre-training Process.}
    \vspace{-15pt}
    \label{tab: datastes}
\end{table}

\subsection{Model Training}
The training process of our \MyMthd{} consists of two main phases: multi-task pre-training and instruction tuning. 
In the Fig.~\ref{fig: data format}, we show the proportions of video-text pairs and image-text pairs in both stages, as well as the chat template of training data. 
%


\myPara{Multi-task Pre-training.}
During the multi-task pre-training phase, we primarily focus on training the three visual projectors. In this phase, we only adjust the parameters of these three projectors while keeping all other parameters frozen. We utilize two types of data: image-text pairs and video-text pairs. For image-text data, we utilize CC-3M \cite{sharma2018conceptual} and RealWorldQA~\cite{grok1.5}, totaling 559K samples.
As for video-text data, we use WebVid-10M~\cite{Bain21}, CLVERER~\cite{yi2019clevrer}, NEXT-QA~\cite{xiao2021next}, Youcook2~\cite{zhou2018towards}, Charades~\cite{sigurdsson2016hollywood}, Charades-Ego~\cite{sigurdsson2018charades}, TGIF~\cite{li2016tgif}, and ShareGPT4Video~\cite{chen2024sharegpt4video}, totaling 2.04M samples.
%
%
The detailed distribution of multi-task pre-training phase is shown in Tab.~\ref{tab: datastes}.

\myPara{Instruction Tuning.}
During the instruction tuning phase, we train the parameters of all three pre-trained projectors, the projector fusion router, and the large language model decoder. 
The weights of the projector fusion router are initialized randomly except the pre-trained BERT~\cite{devlin2018bert}. 
Meanwhile, we keep the parameters of the visual encoder frozen to maintain their stability and consistency.
The instruction data are derived from Oryx~\cite{liu2024oryx}, as detailed in Table~\ref{tab: ft source}. 
Specifically, we integrate comprehensive datasets that include question-answering and video captioning tasks from VideoChatGPT-Plus~\cite{Maaz2024VideoGPT+}, ScanQA~\cite{azuma_2022_CVPR}, ShareGPT4Video~\cite{chen2024sharegpt4video}, and LLaVA-Hound~\cite{zhang2024direct}. To enhance performance on multiple-choice benchmarks, we have also incorporated Cinepile~\cite{rawal2024cinepile}, NextQA~\cite{xiao2021next}, and PerceptionTest~\cite{patraucean2023perception} into our training dataset.

On one hand, current large models often use massive different datasets~\cite{li2024llavaone, ye2024mplugowl3longimagesequenceunderstanding, zhang2024llavanext-video}, and some methods even use private data~\cite{gpt4v, liu2024oryx, openai2024gpt4o}, making it difficult to objectively evaluate the capabilities of model architectures. On the other hand, full-scale multi-task pre-training and instruction tuning require substantial computational resources and time costs. Therefore, to better highlight the advantages stemming from the model architecture rather than the aggregation of large-scale training data, we not only utilize the aforementioned dataset for multi-task pre-training and instruction tuning but also introduce a simplified setup where only the Video-LLaVA dataset (relatively small and has been adopted by many methods) is employed for these stages.

The multi-task pre-training data for Video-LLaVA consist of a subset of 558K LAION-CC-SBU image-text pairs and 702K video-text pairs provided by Valley~\cite{luo2023valley}. For the instruction tuning stage, the data includes 665K image-text instruction pairs from LLaVA1.5~\cite{liu2023llava} and 100K video-text instruction pairs from Video-ChatGPT~\cite{Maaz2023VideoChatGPT}.
Under this setup, we conduct detailed ablation studies to validate the effectiveness of various components of the model.


 Notably, we only use the weighted fusion of multiple projectors for video data. For image inputs, we use the image-based projector only during the process.

\begin{table}[t]
    \centering
    \small
    \vspace{-10pt}
    \setlength{\abovecaptionskip}{2pt}
    \definecolor{lightlightgray}{gray}{0.8}
    \tablestyle{9pt}{1.0}
    \scalebox{0.85}{
    \begin{tabular}{cclcc}
        \toprule
        \textbf{Modality} & \textbf{Task} & \textbf{Dataset} \\
        \midrule
        \multirow{8}{*}{\textbf{Video-Text}} & \multirow{3}{*}{\textbf{Question Answering}} & VideoChatGPT-Plus~\cite{Maaz2024VideoGPT+} \\
                   & &LLaVA-Hound~\cite{zhang2024direct} \\
                   & & ScanQA~\cite{azuma_2022_CVPR}\\   
                   \cmidrule(lr){2-3}
                   & \textbf{Video Caption} &ShareGPT4Video~\cite{chen2024sharegpt4video} \\
                    \cmidrule(lr){2-3}
                   &\multirow{3}{*}{\textbf{Multi-choice QA}} &NEXT-QA~\cite{xiao2021next} \\
                   & & Cinepile~\cite{rawal2024cinepile} \\
                   & &PerceptionTest~\cite{patraucean2023perception} \\

        \bottomrule
    \end{tabular}
    }
    \caption{ Detailed Data Sources of Instruction Tuning.}
    \label{tab: ft source}
\end{table}

\begin{table*}[t]
    \centering
    \small
    \setlength{\abovecaptionskip}{2pt}
    \definecolor{lightlightgray}{gray}{0.8}
    \tablestyle{3pt}{1.0}
    \begin{tabular}{lcccccccccccc}
        \toprule
        \multirow{2}{*}{\textbf{Method}} &  \textbf{Vison} & \textbf{LLM}  & \multicolumn{2}{c}{\textbf{MSVD}} & \multicolumn{2}{c}{\textbf{ActivityNet}} & \multicolumn{6}{c}{\textbf{Video-ChatGPT}} \\
        \cmidrule(lr){4-5} \cmidrule(lr){6-7} \cmidrule(lr){8-13}
        &\textbf{Encoder} & \textbf{Size} & \textbf{Acc.} & \textbf{Score} & \textbf{Acc.} & \textbf{Score} & \textbf{Correctness} & \textbf{Detail} & \textbf{Context} & \textbf{Temporal} & \textbf{Consistency} & \textbf{Avg.} \\
        \midrule
        GPT4-V~\cite{gpt4v} & GPT-4 & - & - & - & 59.5 & - & 4.09 & 3.88&4.37 & 3.94 & 4.02 & 4.06 \\
        \midrule
       
        Video-LLaVA$\dag$~\cite{lin2023video} & ViT-L & 7B
        & 71.8 & 3.9 & 45.3 & 3.3 & - & - & - & - & - & -\\
        LLaMA-VID$\dag$~\cite{li2024llamavid} & CLIP-G & 7B
        & 69.7 & 3.7 & 47.4 & 3.3 & 2.96 & 3.00 & 3.53 & 2.46 & 2.51 & 2.90\\     
        VideoLLaMA2$\dag$~\cite{damonlpsg2024videollama2} & ViT-L & 7B
        & 68.4 & 3.8 & 46.4 & 3.2 & 2.98 & 2.58 & 3.25 & 2.33 & 2.97 & 2.82\\
        \rowcolor[HTML]{EFEFEF} \textbf{\MyMthd{}}$\dag$ & SIGLIP & 7B
        & \textbf{73.4} & \textbf{4.0} &\textbf{48.8} & \textbf{3.5} & \textbf{3.24} & \textbf{2.76} & \textbf{3.51} & \textbf{2.60} & \textbf{3.06} & \textbf{3.03}\\
 \midrule
        FrozenBiLM~\cite{yang2022frozenbilm} & ViT-L & 1.3B
        & 33.8 & - & 25.9 & - & - & - & - & - & - & - \\
        Video-LLaMA~\cite{damonlpsg2023videollama} & CLIP-G & 7B
        & 51.6 & 2.5 & 12.4 & 1.1 & 1.96 & 2.18 & 2.16 & 1.82 & 1.79 & 1.98\\
        LLaMA-Adapter~\cite{zhang2023llamaadapter} & ViT-B & 7B
        & 54.9 & 3.1 & 34.2 & 2.7 & 2.03 & 2.32 & 2.30 & 1.98 & 2.15 & 2.16\\
        VideoChat~\cite{li2023videochat} & ViT-L & 7B 
        & 56.3 & 2.8 & 26.5 & 2.2 & 2.33 & 2.50 & 2.53 & 1.94 & 2.24 & 2.31\\
        Video-ChatGPT~\cite{Maaz2023VideoChatGPT} & ViT-L & 7B
        & 64.9 & 3.3 & 35.2 & 2.7 & 2.50 & 2.57 & 2.69 & 2.16 & 2.20 & 2.42\\
        Chat-UniVi~\cite{jin2023chatunivi} & ViT-L & 7B
        & 65.0 & 3.6 & 45.8 & 3.2 & 2.89 & 2.91 & 3.46 & 2.89 & 2.81 & 2.99\\
        MovieChat~\cite{song2023moviechat} & CLIP-G & 7B 
        & 75.2 & 3.8 & 45.7 & 3.4 & 2.76 & 2.93 & 3.01 & 2.24 & 2.42 & 2.67\\
        VideoChat~\cite{li2023videochat} & CLIP-G & 7B
        & 56.3 & 2.8 & 26.5 & 2.2 & 2.23 & 2.50 & 2.53 & 1.94 & 2.24 & 2.29\\
        BT-Adapter~\cite{liu2024bt} & CLIP-G & 7B
        & 67.7 & 3.7 & 45.7 & 3.2 & 2.68 & 2.69 & 3.27 & 2.34 & 2.46 & 2.20\\
        VideoChat2-HD~\cite{li2024mvbenchcomprehensivemultimodalvideo} & UMT-L & 7B
        & 70.0 & 3.9 & 49.1 & 3.3 & 3.02 & 2.88 & 3.51 & 2.66 & 2.81 & 2.98\\
        VideoLLaMA2~\cite{damonlpsg2024videollama2} & ViT-L & 7B
        & 70.9 & 3.8 & 50.2 & 3.3 & 3.16 & 3.08 & 3.69 & 2.56 & 3.14 & 3.13\\

        Vista-LLaMA~\cite{ma2023vistallama} & CLIP-G & 7B
        & 65.3 & 3.6 & 48.3 & 3.3 & 2.44 & 2.64 & 3.18 & 2.26 & 2.31 & 2.57\\
        ST-LLM~\cite{liu2024st} & BLIP2 & 7B
        & 74.6 & 3.9 & 50.9 & 3.3 & 3.23 & 3.05 & 3.74 & 2.93 & 2.81 & 3.15\\
        PLLaVA~\cite{xu2024pllava} & ViT-L & 7B
        & 76.6 & 4.1 & 56.3 & 3.5 & 3.21 & 2.86 & 3.62 & 2.33 & 2.93 & 2.99\\
        \rowcolor[HTML]{EFEFEF} \textbf{\MyMthd{}} & SIGLIP & 7B
        & \textbf{74.3} & \textbf{4.1} &\textbf{53.4} & \textbf{3.6} & \textbf{3.43} & \textbf{2.95} & \textbf{3.68} & \textbf{2.65} & \textbf{3.24} & \textbf{3.19}\\        
        \bottomrule
    \end{tabular}
    \caption{Results on Video Question-Answering Benchmarks. $\dag$ denotes the use of the same training data as Video-LLaVA~\cite{lin2023video}.}
    \vspace{-5pt}
    \label{tab: VQA}
\end{table*}

\begin{table}[t]
    \centering
    \footnotesize
    \setlength{\abovecaptionskip}{2pt}
    \definecolor{lightlightgray}{gray}{0.8}
    \tablestyle{2.7pt}{1.0}
    \scalebox{1}{
    \begin{tabular}{lccc}
        \toprule
        \textbf{Method} &  \textbf{EgoSchema} & \textbf{MLVU}  & \textbf{VideoMME}\\
        \midrule
        GPT4-V~\cite{gpt4v}& 55.6 & - & 60.7 \\
        GPT4-O~\cite{openai2024gpt4o} & 72.2 & 66.2 & 77.2 \\
        \midrule
        Video-LLaVA$\dag$~\cite{lin2023video} & 38.4 & 47.3 & 40.4\\
        LLaMA-VID$\dag$~\cite{li2024llamavid} & 38.5 & 33.2 & -\\
        VideoLLaMA2$\dag$~\cite{damonlpsg2024videollama2} & 34.6 & 42.9 & 42.7\\
        \rowcolor[HTML]{EFEFEF} \textbf{\MyMthd{}}$\dag$ 
        & \textbf{50.2} & \textbf{55.3} &\textbf{55.7} \\
        \midrule
        Chat-UniVi~\cite{jin2023chatunivi} & - & - & 45.9\\
        VideoChat2-HD~\cite{li2024mvbenchcomprehensivemultimodalvideo} & 54.4 & 47.9 & 54.6\\
        ShareGPT4Video~\cite{chen2024sharegpt4video} & - & 46.4 & 43.6\\
        LLaVA-NeXT-Video~\cite{liu2024llavanext} & 43.9 & - & 46.5\\
        VideoLLaMA2~\cite{damonlpsg2024videollama2} & 51.7 & 48.5 & 46.6\\
        LongVA~\cite{zhang2024long} & - & 56.3 & 54.3\\
        \rowcolor[HTML]{EFEFEF} \textbf{\MyMthd{}}
        & \textbf{59.2} & \textbf{57.5} &\textbf{54.7} \\
        \bottomrule
    \end{tabular}
    }
    \caption{Results on Long Video Understanding Benchmarks. $\dag$ denotes the use of the same training data as Video-LLaVA~\cite{lin2023video}.}
    \label{tab: long video}
\end{table}

\begin{table*}[ht]
  \tablestyle{2.8pt}{1}
  \setlength{\abovecaptionskip}{2pt}
  \footnotesize
  \scalebox{1.0}{
  \begin{tabular}{lccccccccccccccccccccccc} \toprule
    \multirow{2}{*}{\textbf{Method}}  & \multirow{2}{*}{\textbf{AS}} & \multirow{2}{*}{\textbf{AP}} & \multirow{2}{*}{\textbf{AA}} & \multirow{2}{*}{\textbf{FA}} & \multirow{2}{*}{\textbf{UA}} & \multirow{2}{*}{\textbf{OE}} & \multirow{2}{*}{\textbf{OI}} & \multirow{2}{*}{\textbf{OS}} & \multirow{2}{*}{\textbf{MD}} & \multirow{2}{*}{\textbf{AL}} & \multirow{2}{*}{\textbf{ST}} & \multirow{2}{*}{\textbf{AC}} & \multirow{2}{*}{\textbf{MC}} & \multirow{2}{*}{\textbf{MA}} & \multirow{2}{*}{\textbf{SC}} & \multirow{2}{*}{\textbf{FP}} & \multirow{2}{*}{\textbf{CO}} & \multirow{2}{*}{\textbf{EN}} & \multirow{2}{*}{\textbf{ER}} & \multirow{2}{*}{\textbf{CI}} & \multirow{2}{*}{\textbf{Avg.}} \\ 
    \\ \midrule
     GPT-4V~\cite{gpt4v} 
      & 55.5 & 63.5 & 72.0 & 46.5 & 73.5 & 18.5 & 59.0 & 29.5 & 12.0 & 40.5 & 83.5 & 39.0 & 12.0 & 22.5 & 45.0 & 47.5 & 52.0 & 31.0 & 59.0 & 11.0 & 43.5\\ \midrule
        VideoLLaMA2$\dag$~\cite{damonlpsg2024videollama2}
     & 59.5 & 46.5 & 64.5 & 45.4 & 58.6 & 47.7 & 48.0 & 37.3 & 23.5 & 31.0 & 75.0 & 40.5 & 32.5 & 46.0 & 38.0 & 36.5 & 49.0 & 27.5 & 43.5 & 38.5 & 44.5 \\
          LLaMA-VID$\dag$~\cite{li2024llamavid} 
     & - & - & - & - & - & - & - & - & -& - & - & - & - & - & - & - & - & - & - & - & 41.9\\
     
    \rowcolor[HTML]{EFEFEF} \textbf{\MyMthd{}}$\dag$
     & \textbf{58.9} & \textbf{51.3}  & \textbf{75.4}  & \textbf{47.6}  & \textbf{73.0}  & \textbf{57.1}  & \textbf{66.5}  & \textbf{36.0}  & \textbf{19.4}  & \textbf{47.8}  & \textbf{90.0}  & \textbf{48.5}  & \textbf{32.0}  & \textbf{52.5}  & \textbf{46.5}  & \textbf{44.0}  & \textbf{63.0}  & \textbf{30.5}  & \textbf{54.0}  & \textbf{38.5}  & \textbf{51.7}  \\
     \midrule
     Video-LLaMA~\cite{damonlpsg2023videollama}  & 27.5 & 25.5 & 51.0 & 29.0 & 39.0 & 48.0 & 40.5 & 38.0 & 22.5 & 22.5 & 43.0 & 34.0 & 22.5 & 32.5 & 45.5 & 32.5 & 40.0 & 30.0 & 21.0 & 37.0 &34.1\\
     LLaMA-Adapter~\cite{zhang2023llamaadapter} 
     & 23.0 & 28.0 & 51.0 & 30.0 & 33.0 & 53.5 & 32.5 & 33.5 & 25.5 & 21.5 & 30.5 & 29.0 & 22.5 & 41.5 & 39.5 & 25.0 & 31.5 & 22.5 & 28.0 & 32.0 & 31.7\\
     Video-ChatGPT~\cite{Maaz2023VideoChatGPT}  
     & 23.5 & 26.0 & 62.0 & 22.5 & 26.5 & 54.0 & 28.0 & 40.0 & 23.0 & 20.0 & 31.0 & 30.5 & 25.5 & 39.5 & 48.5 & 29.0 & 33.0 & 29.5 & 26.0 & 35.5 & 32.7\\
     VideoChat~\cite{li2023videochat} 
     & 33.5 & 26.5 & 56.0 & 33.5 & 40.5 & 53.0 & 40.5 & 30.0 & 25.5 & 27.0 & 48.5 & 35.0 & 20.5 & 42.5 & 46.0 & 26.5 & 41.0 & 23.5 & 23.5 & 36.0 & 35.5\\
     VideoChat2-HD~\cite{li2024mvbenchcomprehensivemultimodalvideo}
     & 66.0 & 47.5 & 83.5 & 49.5 & 60.0 & 58.0 & 71.5 & 42.5 & 23.0 & 23.0 & 88.5 & 39.0 & 42.0 & 58.5 & 44.0 & 49.0 & 36.5 & 35.0 & 40.5 & 65.5 & 51.1\\
     ST-LLM~\cite{liu2024st}
     & 66.0 & 53.5 & 84.0 & 44.0 & 58.5 & 80.5 & 73.5 & 38.5 & 42.5 & 31.0 & 86.5 & 36.5 & 56.5 & 78.5 & 43.0 & 44.5 & 46.5 & 34.5 & 41.5 & 58.5 & 54.9\\
     PLLaVA~\cite{xu2024pllava}
     & 58.0 & 49.0 & 55.5 & 41.0 & 61.0 & 56.0 & 61.0 & 36.0 & 23.5 & 26.0 & 82.0 & 39.5 & 42.0 & 52.0 & 45.0 & 42.0 & 53.5 & 30.5 & 48.0 & 31.0 & 46.6\\
     VideoLLaMB~\cite{videollamb}
     & 54.5 & 47.0 & 86.5 & 44.5 & 52.0 & 79.0 & 58.5 & 32.0 & 47.0 & 33.0 & 82.5 & 40.5 & 52.0 & 82.0 & 40.5 & 37.5 & 43.0 & 31.0 & 42.5 & 60.0 & 52.5 \\
     VideoLLaMA2~\cite{damonlpsg2024videollama2}
     & - & - & - & - & - & - & - & - & -& - & - & - & - & - & - & - & - & - & - & - & 54.6\\


    \rowcolor[HTML]{EFEFEF} \textbf{\MyMthd{}}
     & \textbf{71.4} & \textbf{63.2}  & \textbf{80.8}  & \textbf{51.2}  & \textbf{78.1}  & \textbf{92.4}  & \textbf{78.5}  & \textbf{39.5}  & \textbf{62.7}  & \textbf{54.5}  & \textbf{95.5}  & \textbf{53.5}  & \textbf{78.5}  & \textbf{91.0}  & \textbf{67.0}  & \textbf{50.5}  & \textbf{74.0}  & \textbf{35.0}  & \textbf{57.0}  & \textbf{64.5}  & \textbf{66.9}  \\
    \bottomrule
  \end{tabular}
  }
    \caption{Results on MVBench. $\dag$ denotes the use of the same training data as Video-LLaVA~\cite{lin2023video}.
  }\label{tab: mvbench}
  \vspace{-5pt}
\end{table*}

\begin{table}[t]
      \centering
      \setlength{\abovecaptionskip}{4pt}
      \tablestyle{5pt}{1}

    \begin{tabular}{ccccccc} \toprule
    \textbf{Image-} & \textbf{Spatial-} & \textbf{Token-} & \multirow{2}{*}{\textbf{MVBench}} & \multirow{2}{*}{\textbf{VideoMME}}\\
     \textbf{based}  &\textbf{temporal}  & \textbf{compress}  \\
    \midrule
    \cmark &        &        & 48.6 &50.5\\
           & \cmark &        & 49.1 &52.1\\
           &        & \cmark & 45.8 &51.3\\
    \cmark &        & \cmark & 50.4 &53.4\\
    \cmark & \cmark &        & 51.3 &53.9\\
           & \cmark & \cmark & 50.7  &54.2\\
    \cmark & \cmark & \cmark & \textbf{51.7}&\textbf{55.7}\\
    \bottomrule
    \end{tabular}
        \caption{
    Ablation study on the effectiveness of different type of visual projectors.
    }
    \label{tab:components}
    \vspace{-10pt}
\end{table}

\section{Experiments}
\label{sec:exp}
\subsection{Experimental Setup}

\myPara{Implementation Details.} We employ the Qwen2.5-7B-Instruct~\cite{qwen2.5} as the LLM and SigLIP (so400m-patch14-384)~\cite{zhai2023sigmoid} as the visual encoder. 
All experiments are performed on 8 NVIDIA A100 GPUs.

\subsection{Main Results}

\myPara{Results on Video Question Answering Benchmark.}
In \tabref{tab: VQA}, we demonstrate the performance of our \MyMthd{} against state-of-the-art methods on three zero-shot video QA benchmarks. MSVD-QA~\cite{chen2011collecting} is a dataset comprising questions about short real-world video clips, typically lasting 10-15 seconds. ActivityNet-QA~\cite{caba2015activitynet} consists of human-annotated action-related QA pairs derived from the ActivityNet dataset, with an average duration of 2 minutes. Additionally, we evaluate our model on 
VideoChatGPT~\cite{Maaz2023VideoChatGPT} benchmark, which assesses five key aspects of video understanding: correctness of information, detail orientation, context understanding, temporal understanding, and consistency.

\myPara{Results on Long Video Understanding Benchmark.}
To demonstrate that our method can handle various video scenarios, we present several relatively long video understanding benchmarks in \tabref{tab: long video}.
Among these, EgoSchema~\cite{mangalam2024egoschema} consists of egocentric videos with an average duration of 180 seconds.
MLVU~\cite{zhou2024mlvu} focuses on long video understanding, with video lengths ranging from 3 to 120 minutes.
VideoMME~\cite{fu2024videomme}, containing diverse video domains and durations (ranging from minutes to hours), is a relatively comprehensive video understanding benchmark.

\myPara{Results on MVBench.}
Besides the VQA benchmarks mentioned above, we also conduct experiments on MVBench~\cite{li2024mvbenchcomprehensivemultimodalvideo}, a comprehensive video understanding benchmark covering 20 tasks organized in the form of multiple-choice questions in \tabref{tab: mvbench}.
\MyMthd{} achieves state-of-the-art (SOTA) performance in almost all tasks, demonstrating that our instruction-driven adaptive projector fusion strategy effectively leverages the strengths of different projectors and overcomes the limitations of a single projector in specific domains.

    

\subsection{Ablation Study}

\myPara{Effectiveness of Each Projectors.} To demonstrate the impact of different projectors, in Tab.~\ref{tab:components}, we first conduct ablation studies using various numbers and types of projectors.
We aware that 
the tokens derived from the image-based and spatial-temporal projectors are temporally and spatially alignable. However, the tokens generated by the token-compress projector, due to the token compression paradigm, cannot be directly aligned in terms of temporal and spatial dimensions with those from the other two projectors. The features resulting from the token-compress projector, when added to those from the other two projectors, can to some extent disrupt the spatial and temporal relationships. Nevertheless, incorporating the tokens from the token-compress projector significantly preserves the temporal integrity, as experimental results have demonstrated that this temporal integrity brings substantial benefits.
It can be observed that compared to using a single type of projector, each addition of a new type of projector results in performance improvements on both MVBench and VideoMME benchmark.


 \begin{table}[t]
\small
\centering
      \setlength{\abovecaptionskip}{4pt}
  \tablestyle{10pt}{1}

  \begin{tabular}{lcccccc} 
    \toprule
    \textbf{Method} & \textbf{MVBench}  & \textbf{VideoMME}   \\
    \midrule
    Average& 50.4 & 53.6 \\
    Concat &51.2 & 54.8\\
    Random weights &50.1 & 52.9\\ 
    Random choose& 50.9 & 53.4 \\ 
    Projector Fusion Router&  \textbf{51.7} & \textbf{55.7} \\ 
    \bottomrule
  \end{tabular}
        \caption{Ablation study on the projector fusion strategies.
  }
  \vspace{-10pt}
  \label{tab:select}
\end{table}

\myPara{Imapct of Different Projector Fusion Strategies. }
To demonstrate the effectiveness of our proposed instruction-driven adaptive router, we conduct ablation studies using different projector fusion strategies in Tab.~\ref{tab:select}. Specifically, we perform experiments under average, concatenation, random weight, and random choose settings, in addition to our proposed method.
.
The results show that our adaptive router outperforms other strategies on both the MVBench and Video MME benchmarks. This demonstrates that our projector fusion router can effectively determine the weight of each projector's contribution to the final visual embedding based on user instructions, thereby better adapting to different task scenarios.

\myPara{Stacked Projectors v.s. Different Types of Projectors.} A potential concern regarding \MyMthd{} might be that the performance improvements are not due to the complementarity of multiple types of projectors, but rather the sheer number of projectors. To address this, we conducted a comparative experiment where we retained the instruction-driven adaptive router but replaced the different types of projectors with repeatedly stacked projectors of the same type, as shown in Tab. \ref{tab: stacked}. The results indicate that using repeated projectors offers some improvement over employing a single projector but still falls short when compared to utilizing different types of projectors. This demonstrates that the information extracted by different projectors is complementary. Our instruction-driven adaptive router can leverage the strengths of diverse projectors effectively, thereby enhancing the overall performance.
\begin{figure*}[t]
    \centering
    \vspace{-0.5cm}
    \includegraphics[width=0.98\linewidth]{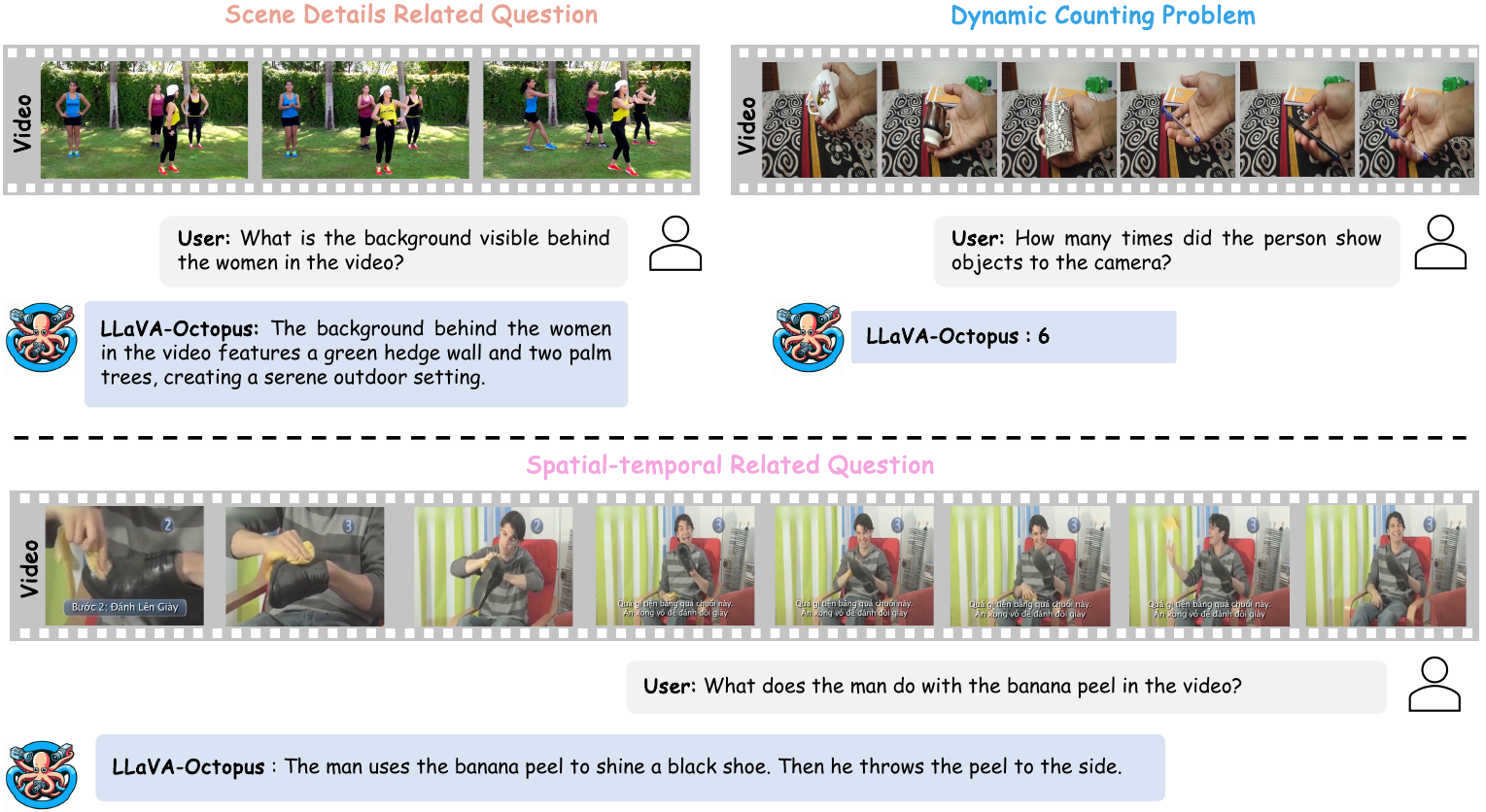}
    
    \caption{Comparison of Various Projector Fusion Strategies.
    \vspace{-0.3cm}}
    \label{fig: example} 
\end{figure*}
 \begin{table}[t]
\small
\centering
      \setlength{\abovecaptionskip}{4pt}
  \tablestyle{7pt}{1}

  \begin{tabular}{llccccc} 
    \toprule
    \textbf{Projector} &\textbf{Method} & \textbf{MVBench}& \textbf{VideoMME}  \\
    \midrule
    \multirow{2}{*}{\textbf{Image-based}}& Single  & 48.6 &50.5\\
    &Stacked & 49.2 & 51.0 \\
    \multirow{2}{*}{\textbf{Spatial-temporal}}& Single  & 49.1 &52.1\\
    &Stacked & 49.3 & 52.1 \\
    \multirow{2}{*}{\textbf{Token-compress}}& Single  & 45.8 &51.3\\
    &Stacked & 46.7 & 51.6  \\
    \midrule
    \textbf{All}  &Fusion & \textbf{50.7}  &\textbf{54.2}\\
    \bottomrule
  \end{tabular}
        \caption{ Ablation study on repeatedly stacked same projectors and different projectors.
  }
  \vspace{-15pt}
  \label{tab: stacked}
\end{table}

\subsection{Discussion on projectors' weight.}
As shown in \tabref{tab:select} and Tab. \ref{tab: stacked}, \MyMthd{} outperforms other approaches that utilize different projector fusion strategies or projector types, indicating the projector fusion strategy is not trival. 
Therefor, we further analyze the specific weight values assigned to the projectors.  
However, the lack of explicit categorization in current benchmarks makes systematic per-task statistics impractical. Consequently, we examine the projectors' weights from some examples (Fig. \textcolor{iccvblue}{1}, Fig. \textcolor{iccvblue}{2}, Fig. \textcolor{iccvblue}{3}
in supplementary materials ). For 5 scene detail cases, the image-based projector dominates (\textbf{avg. weight=0.71}); for 2 spatial-temporal cases, the spatial-temporal projector prevails (\textbf{avg. weight=0.76}); and for 5 dynamic counting cases, the token-compress projector is prioritized (\textbf{avg. weight=0.61}). 

These findings demonstrate that our instruction-driven adaptive router adaptively emphasizes task-relevant projectors. The weight distribution across different tasks also highlights the complementarity of the features extracted by different projectors, reinforcing the effectiveness of our approach in leveraging their respective strengths.

\subsection{Qualitative analysis}
In Fig.~\ref{fig: example}, we demonstrate some qualitative examples of \MyMthd{}
\MyMthd{} achieves correct responses in each of these scenarios, illustrating its ability to integrate the strengths of different visual projectors and overcome the inherent limitations imposed by a single projector. This versatility allows our method to perform well not only on specific types of problems but also in a wide range of comprehensive instruction scenarios.

\subsection{Ablations on Different Vison Encoders}
Since the Vision Encoder is a crucial component of the MLLM and is directly connected to the visual projector, the quality of visual features significantly impacts the performance of the MLLM.
Therefore, we have empirically compared two prevalent visual encoders, CLIP and SigLIP (the two most common visual encoders in MLLM), to ensure robustness even though our core contribution lies in the instruction-driven adaptive projector rather than in the design of the visual encoder. As shown in Table 6, SigLIP consistently outperformed CLIP on both MVBench and VideoMME benchmarks. We therefore adopt SigLIP as the default encoder.

 \begin{table}[h]
\small
\centering
      \setlength{\abovecaptionskip}{4pt}
  \tablestyle{12pt}{1}

  \begin{tabular}{lcccccc} 
    \toprule
    \textbf{Vision Encoder} & \textbf{MVBench}  & \textbf{VideoMME}   \\
    \midrule
CLIP & 62.6 & 50.8 \\
SigLIP & \textbf{66.9} (+4.3\%) & \textbf{54.7} (+3.9\%) \\
    \bottomrule
  \end{tabular}
        \caption{Ablation study on different vision encoders.
  }
  \vspace{-10pt}
  \label{tab:encoder}
\end{table}

\subsection{Selection of visual projectors}
Our work’s innovation centers on the instruction-driven adaptive router, not on claiming the superiority of specific projectors. Therefore, the projectors we select are all widely adopted in MLLMs for reproducibility. Since our fusion mechanism is architecture-agnostic, here we conduct an experiment of replacing the token-compress projectors from LLaMA-VID~\cite{li2024llamavid} projector to PLLaVA’s~\cite{xu2024pllava} projector in \tabref{tab: projector}.
It can be seen that using PLLaVA’s adaptive pooling projector even brings some improvements on performance (67.4 on MVBench and 56.1 on VideoMME), proving the adaptability of \MyMthd{}. 
We believe that some specific-designed architectures for visual projector would further improve the model performance and regard this as a promising direction for future research.


 \begin{table}[h]
\small
\centering
      \setlength{\abovecaptionskip}{4pt}
  \tablestyle{7.5pt}{1}

  \begin{tabular}{lcccccc} 
    \toprule
    \textbf{Visual Projector} & \textbf{MVBench}  & \textbf{VideoMME}   \\
    \midrule
LLaMA-VID~\cite{li2024llamavid} & 66.9 & 54.7 \\
PLLaVA~\cite{xu2024pllava} & 67.4 (+0.5\%) & 56.1 (+1.4\%) \\
    \bottomrule
  \end{tabular}
        \caption{Ablation study on different visual projector.
  }
  \vspace{-10pt}
  \label{tab: projector}
\end{table}

\section{Conclusions}
In this paper, we introduce \MyMthd{}, a novel video multimodal large language model. \MyMthd{} dynamically fuses the visual embedding from different visual projectors via an instruction-driven adaptive router, effectively leveraging the unique strengths of each projector. By dynamically combining the most suitable features, \MyMthd{} significantly enhances its performance in various multimodal tasks. Our experimental results demonstrate that \MyMthd{} achieves outstanding performance across multiple benchmarks,
highlighting its promising application potential.
\newpage
\appendix
\setcounter{page}{1}

\section{More Discussions}

\section{More Comparisons of Different Projectors}
\begin{figure*}[t]
    \vspace{-5mm}
    \centering
    \includegraphics[width=0.98\linewidth]{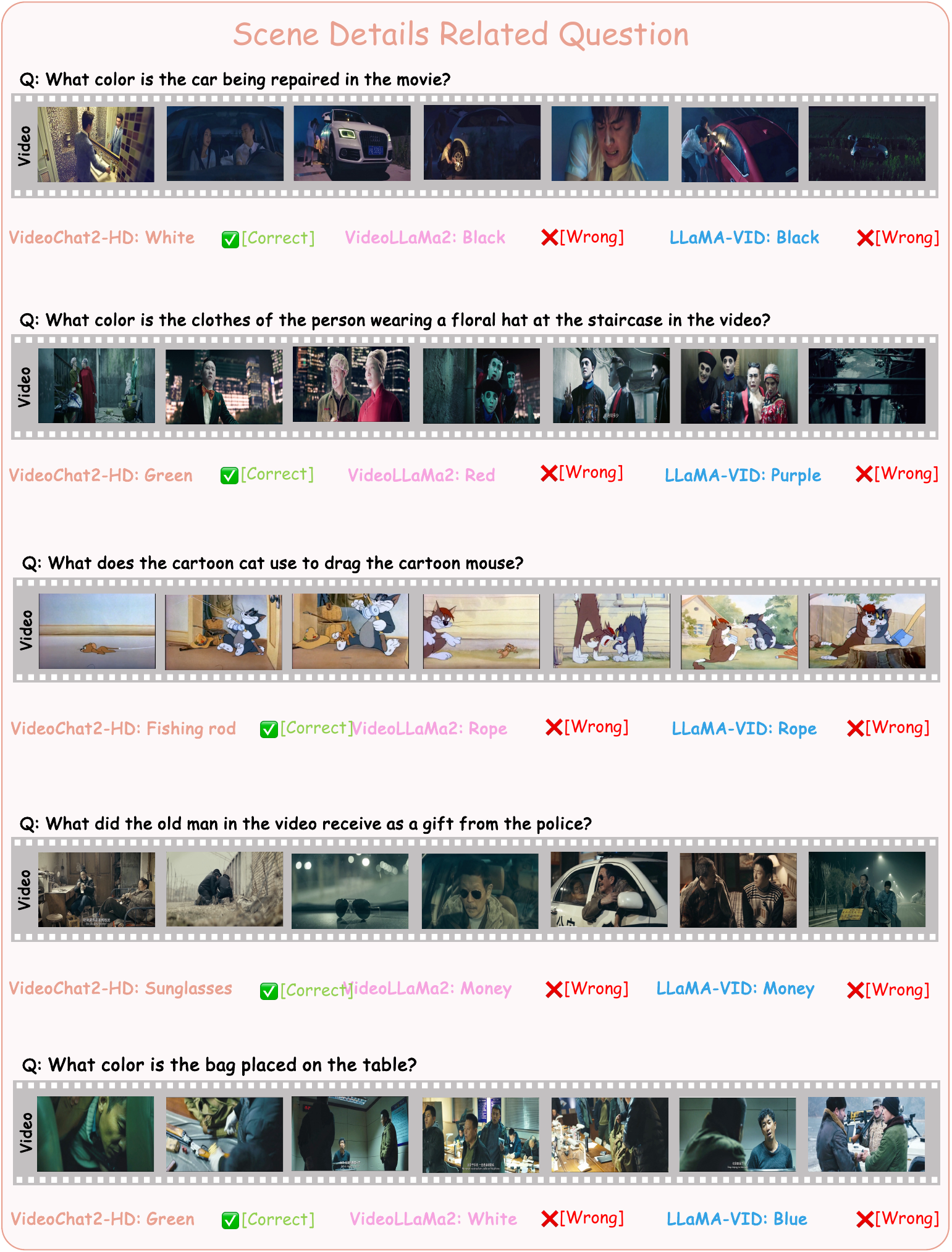}
    \vspace{-0.1cm}
    \caption{\textbf{More examples of scene details related question.}\vspace{-0.3cm}}
    \label{fig: detail} 
\end{figure*}

\begin{figure*}[t]
    \vspace{-5mm}
    \centering
    \includegraphics[width=0.98\linewidth]{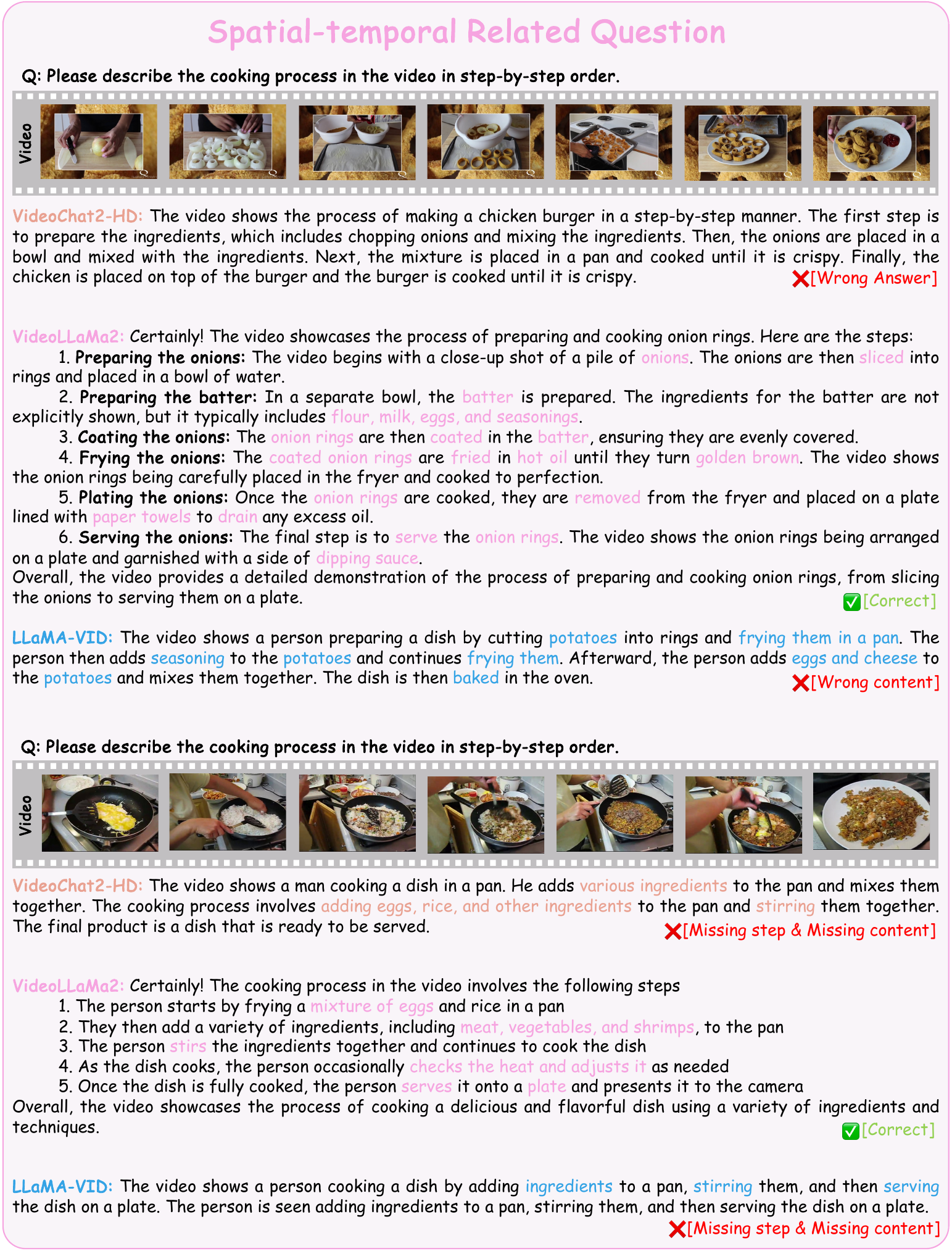}
    \vspace{-0.1cm}
    \caption{\textbf{More examples of spatial-temporal related question.}\vspace{-0.3cm}}
    \label{fig:stc} 
\end{figure*}

\begin{figure*}[t]
    \vspace{-5mm}
    \centering
    \includegraphics[width=0.98\linewidth]{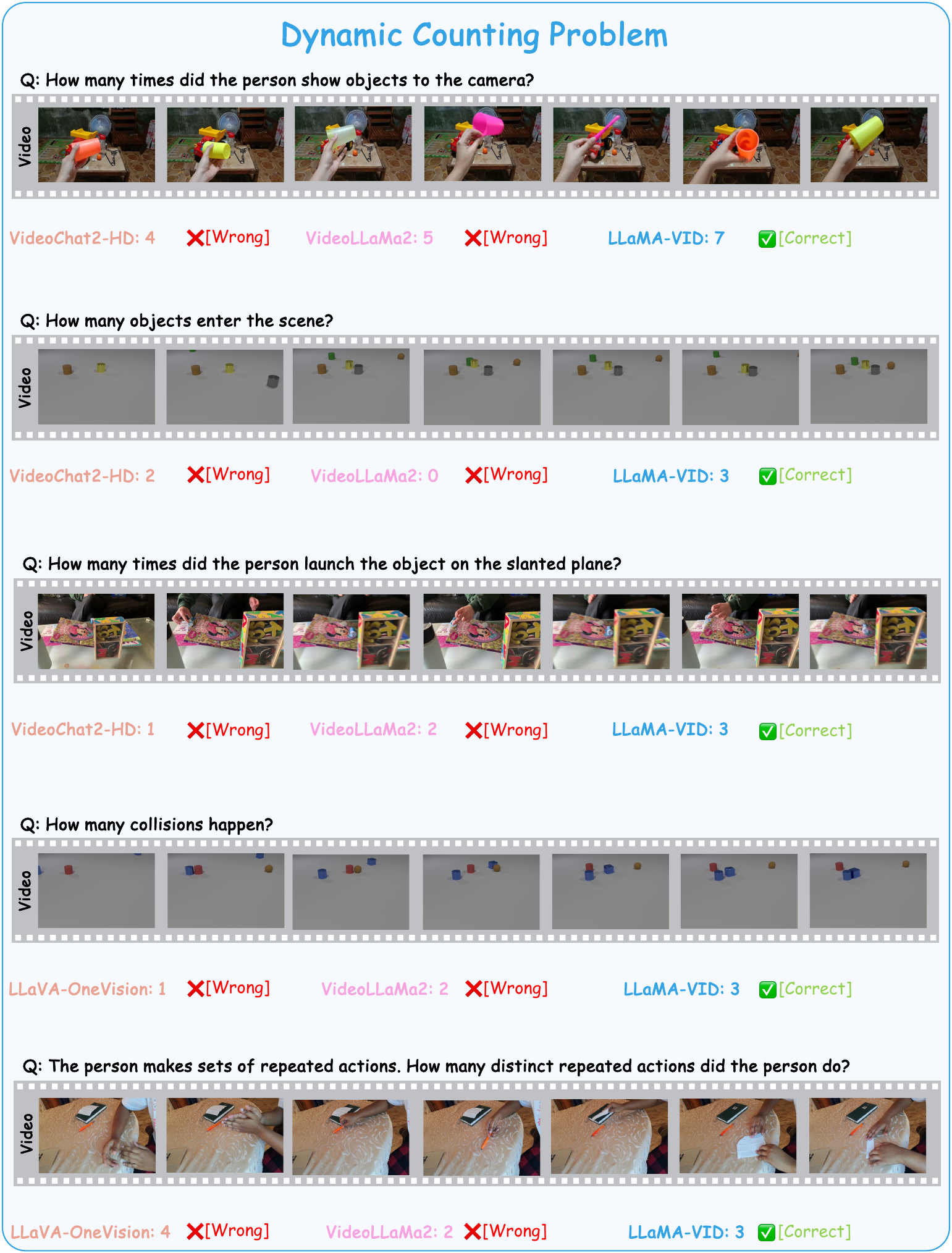}
    
    \vspace{-0.1cm}
    \caption{\textbf{More examples of dynamic counting question.}\vspace{-0.3cm}}
    \label{fig:counting} 
\end{figure*}

As discussed in Sec. \textcolor{iccvblue}{1} of our main paper, the significance of visual projectors and the applicability of different types of visual projectors to various visual task scenarios constitute a crucial motivation for \MyMthd{}. We have provided some examples in Fig.~\textcolor{iccvblue}{2} of our main paper to illustrate this phenomenon. To further demonstrate its generalizability and reinforce our motivation, we supplement more additional examples in Fig. \ref{fig: detail}, Fig. \ref{fig:stc} and Fig. \ref{fig:counting}.

Specifically, in Fig.~\ref{fig: detail}, we present examples of Scene Details Related Questions using representative methods of the three types of projectors. In complex backgrounds, when questions require a more detailed understanding of the scene, the Image-based Projector demonstrates superior performance. In Fig.~\ref{fig:stc}, we show examples of Spatial-temporal Related Questions using representative methods of the three types of projectors. It can be seen that the method based on the Spatial-temporal Projector, VideoLLaMA2~\cite{damonlpsg2024videollama2}, shows a clear advantage. In Fig.~\ref{fig:counting}, we demonstrate the effectiveness of different projector methods in problems that require temporal consistency. Similar to the discussion in the paper, we chose Dynamic Counting Problems to represent this category. It is evident that the temporal consistency of both the Image-based Projector and the Spatial-temporal Projector is severely compromised, leading to poor performance in this type of problem.  In contrast, the method with Token-compress Projector shows good performance in this category.


\section{More  Qualitative Results}
\begin{figure*}[t]
    \vspace{-5mm}
    \centering
    \includegraphics[width=0.98\linewidth]{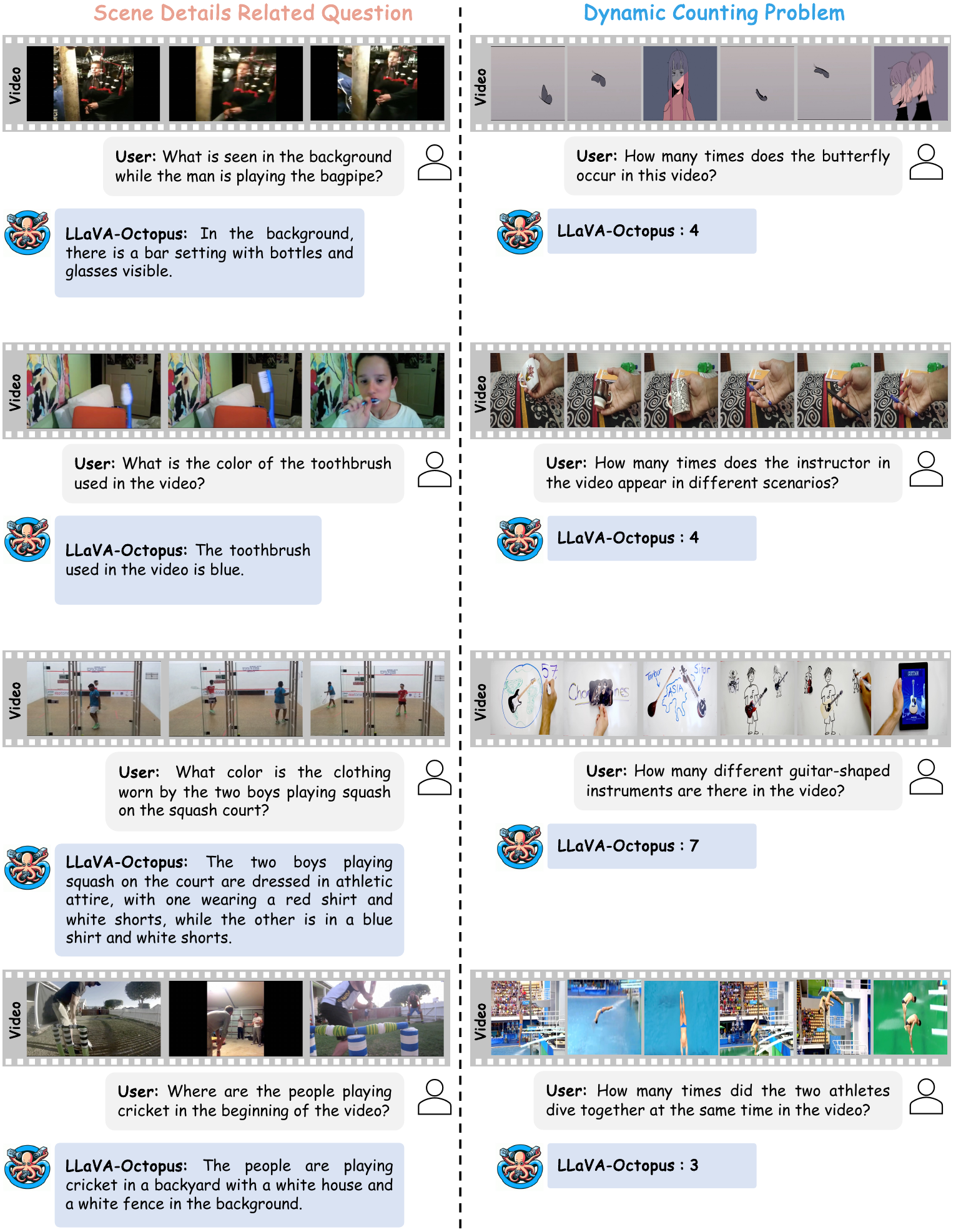}
    \vspace{-0.1cm}
    \caption{\textbf{More qualitative results of \MyMthd{}.}\vspace{-0.3cm}}
    \label{fig:sample_1} 
\end{figure*}

\begin{figure*}[t]
    \vspace{-5mm}
    \centering
    \includegraphics[width=0.98\linewidth]{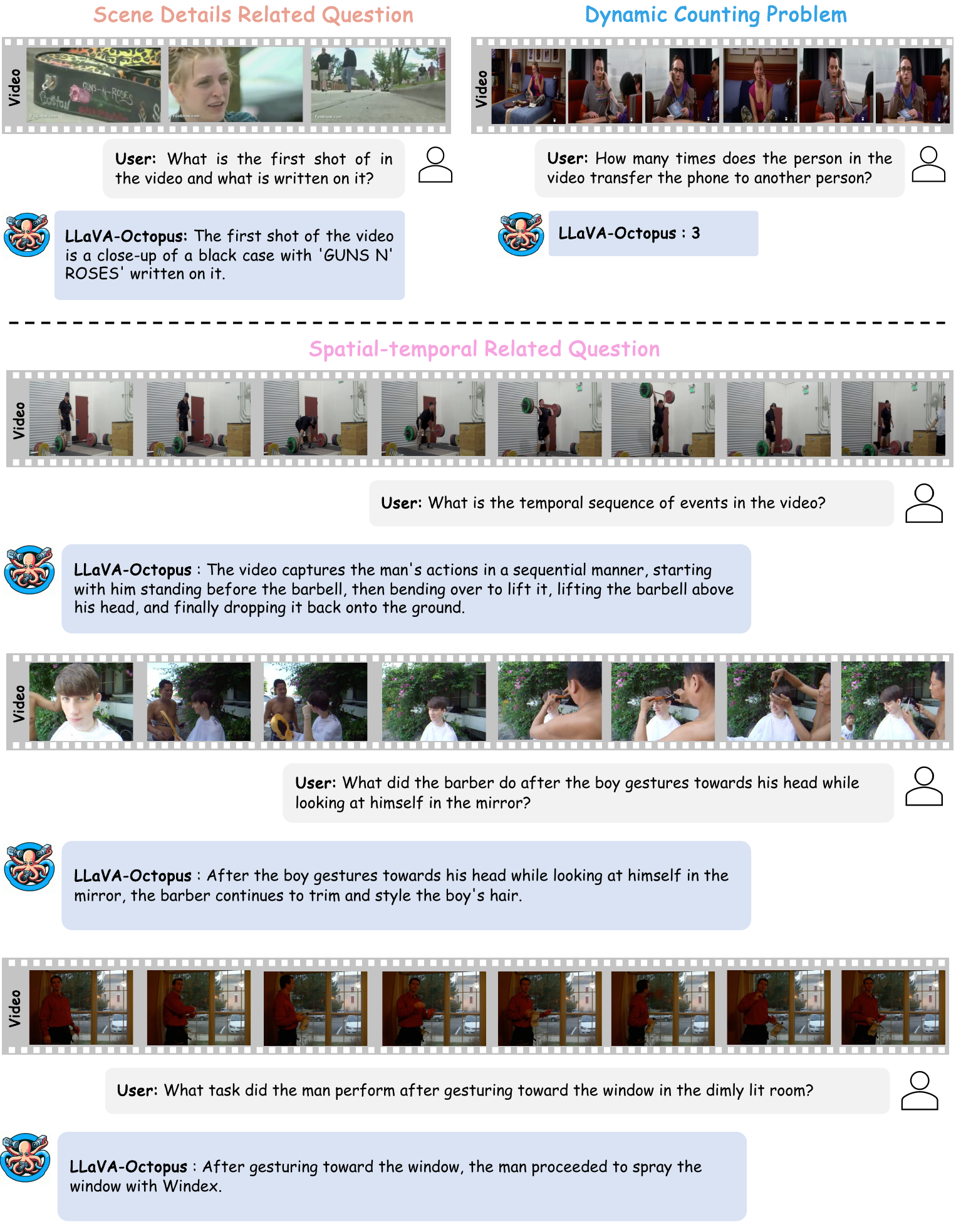}
    \vspace{-0.1cm}
    \caption{\textbf{More qualitative results of \MyMthd{}.}\vspace{-0.3cm}}
    \label{fig:sample_2} 
\end{figure*}

In our main paper, we claim that proposed \MyMthd{} can tackle different video understanding scenarios and comprehensive user instructions. We have verified this through both extensive quantitative and qualitative experiments in Sec.\textcolor{iccvblue}{4} of our main paper. Here, we present more qualitative results in \figref{fig:sample_1} and \figref{fig:sample_2} to further support our conclusion.

Specifically, in \figref{fig:sample_1} and \figref{fig:sample_2}, we present the performance of our LLaVA-Octopus on three types of questions: Scene Details Related Questions, Spatial-temporal Related Questions, and Dynamic Counting Problems. It can be seen that due to the reasonable integration of image-based projector, spatial-temporal projector, and token-compress projector in our model architecture, our LLaVA-Octopus is capable of providing accurate answers to all three types of questions.
\newpage


\clearpage

{
    \small
    \bibliographystyle{ieeenat_fullname}
    \bibliography{main}
}


\end{document}